\documentclass{article}
\usepackage{spconf,amsmath,amssymb,graphicx}
\usepackage{cite,booktabs,multirow}
\usepackage[T1]{fontenc}
\usepackage[hyphens]{url}
\usepackage{hyperref}

\title{COMPLEMENTARY rPPG-DERIVED AND LIP-REGION FREQUENCY CUES FOR TALKING-FACE DEEPFAKE DETECTION}

\twoauthors
  {Othmane Harraq}
  {Temple University\\
   Philadelphia, PA, USA\\
   othmane.harraq@temple.edu}
  {Tamer Aldwairi}
  {Temple University\\
   Philadelphia, PA, USA\\
   aldwairi@temple.edu}

\begin{document}

\maketitle
\begin{abstract}
Talking-face (TF) deepfakes are detected unevenly by rPPG-based methods across generators. We study two lightweight visual-only cues, rPPG-derived waveforms extracted by RhythmFormer and lip-region discrete cosine transform (DCT) coefficients, on the seven TF methods of Celeb-DF++ under a subject-independent protocol. In-domain, lip-region DCT matches or exceeds the rPPG-derived 1D ResNet on every method except SadTalker, and Concat fusion reaches AUC 0.891 against 0.824 and 0.827 for the unimodal baselines. Under leave-one-generator-out evaluation the cues split: each transfers clearly better to three held-out methods, and IP-LAP is near chance for both. Concat averages 0.798 but falls below rPPG alone where DCT transfers poorly, so static fusion only partly exploits this complementarity. Lip-region DCT outperforms full-face DCT on six of seven methods. We treat the rPPG-derived signal as an empirical cue and do not claim it is cardiac in origin.
\end{abstract}
\section{Introduction}
\label{sec:intro}

Talking-face (TF) deepfake synthesis animates a source face with audio to produce photorealistic video. Remote photoplethysmography (rPPG) recovers pulse-related signals from periodic skin color variations \cite{verkruysse2008rppg}, and has been applied to deepfake forensics since FakeCatcher \cite{ciftci2020fakecatcher}, with later detectors building on heart-rate estimation \cite{hernandez2020deepfakeson} and attentional rhythm analysis \cite{qi2020deeprhythm}. These methods were developed on face-swap forgeries, which retain the source video's skin-color dynamics beneath the swapped region. TF synthesis instead regenerates most or all of the face, so real temporal skin-color variation is absent or only partially retained. In this setting, prior work \cite{harraq2027paper1} showed that rPPG-derived temporal signals separate real from TF video (AUC 0.826) without establishing a cardiac origin, while [5] measured DeepFakesON-Phys \cite{hernandez2020deepfakeson} at AUC 0.576 on TF forgeries. Existing TF detectors either require synchronized audio \cite{liu2024avlip} or learn mouth-region cues from visual input \cite{haliassos2021lips, datta2025lipsync}.

Our contributions are:
\begin{itemize}
\item An analysis of method-dependent complementarity between rPPG-derived temporal signals and lip-region DCT features across all seven Celeb-DF++ TF methods under a subject-independent protocol.
\item A comparison of three fusion architectures against both unimodal baselines, in-domain and under leave-one-generator-out evaluation with identity-disjoint samples, where the cues split by held-out method and static fusion trails the better single cue on three of seven.
\item Evidence that lip-region DCT outperforms full-face DCT on six of seven methods.
\end{itemize}

Our code and results are available at:
\begin{center}
\small\url{https://github.com/AI-Advanced-Vision-Forensics-Lab/rppg-dct-fusion}
\end{center}

\section{Related Work}
\label{sec:related}

\textbf{Talking-face synthesis.} TF methods drive a source face with audio. SadTalker \cite{zhang2023sadtalker} estimates 3D morphable model coefficients and renders head motion via learned motion fields. IP-LAP \cite{zhong2023iplap} re-renders the lower face from audio-driven landmarks and reference appearance priors. AniTalker \cite{liu2024anitalker}, EDTalk \cite{tan2024edtalk}, Real3D-Portrait \cite{ye2024real3d}, EchoMimic \cite{chen2025echomimic}, and FLOAT \cite{ki2025float} (flow matching in a learned motion latent) represent recent audio-driven portrait animation approaches.

\textbf{rPPG-based detection.} rPPG recovers cardiac pulse waveforms from periodic skin color variations \cite{verkruysse2008rppg}. RhythmFormer \cite{zou2025rhythmformer} achieves state-of-the-art extraction via periodic sparse attention. FakeCatcher \cite{ciftci2020fakecatcher} pioneered rPPG forensics; subsequent work \cite{hernandez2020deepfakeson, qi2020deeprhythm} focused exclusively on face-swap data. Prior work \cite{harraq2027paper1} reported, to our knowledge, the first systematic evaluation of rPPG-derived signals for TF detection, and cautioned that their discriminative content may not be cardiac, echoing concerns raised for face-swap data \cite{damelio2023cautionary}. Relative to \cite{harraq2027paper1}, which evaluated rPPG-derived signals alone, we add lip-region DCT features, compare three fusion architectures, and extend its leave-one-generator-out evaluation to the lip-region cue and to fusion.

\textbf{Frequency-domain forensics.} Frequency-aware detectors such as F3-Net \cite{qian2020thinking} exploit DCT-domain artifacts left by synthesis pipelines. We apply DCT to the lip region, where region-level re-rendering is expected to leave artifacts.

\section{Methodology}
\label{sec:method}

\subsection{rPPG Waveform Extraction}
We extract rPPG waveforms using RhythmFormer \cite{zou2025rhythmformer} with the UBFC\_cross checkpoint. MediaPipe BlazeFace \cite{bazarevsky2019blazeface} detects faces with $1.5\times$ bounding box expansion; frames are resized to $128\times128$ and globally z-score normalized. Real videos yield stride-60 sliding windows (579 usable source videos, 2,371 real windows); fake videos yield one waveform each (first 160 frames or linearly interpolated if $60 \le T < 160$). Following \cite{harraq2027paper1}, we call this signal rPPG-derived and do not claim its discriminative content is cardiac. Per-window z-score normalization reduces identity-specific amplitude:
\begin{equation}
\tilde{\mathbf{x}} = \frac{\mathbf{x} - \mu_{\mathbf{x}}}
{\sigma_{\mathbf{x}} + \epsilon}, \quad \epsilon = 10^{-6}.
\end{equation}

\subsection{Lip-Region DCT Features}
MediaPipe FaceLandmarker \cite{grishchenko2020attention} identifies lip landmarks (indices 61, 291, 0, 17); a $64\times64$ grayscale crop with 20px padding is passed through a 2D DCT. The top-left $8\times8$ block in zigzag order yields a 64-dimensional feature vector capturing low-frequency lip texture. The DCT vector is computed from a single frame: the center frame of each real window and the middle frame of each fake video. The lip region is motivated by methods that re-render the mouth, which may leave frequency artifacts at the region boundary.

\subsection{Fusion Architectures}

We compare two unimodal baselines and three fusion architectures.

\textbf{rPPG-only (1D ResNet).} A stem convolution, three residual stages with configuration $(2,2,2)$ and channel widths 32, 64, 128, global average pooling, and a linear classifier (240K parameters, as in \cite{harraq2027paper1}).

\textbf{DCT-only (MLP).} A two-layer network applies Linear$(64\to64)$ followed by batch normalization, ReLU activation, and dropout, then projects to a scalar logit via Linear$(64\to1)$.

\textbf{Gated fusion} \cite{arevalo2017gated}\textbf{.} A linear layer on $[\mathbf{h}_\text{rPPG};\mathbf{h}_\text{DCT}]$ followed by a sigmoid produces a per-sample weight $\alpha \in (0,1)$:
\begin{equation}
\mathbf{h} = \alpha\mathbf{h}_\text{rPPG} + (1-\alpha)\mathbf{h}_\text{DCT}.
\end{equation}

The fused $\mathbf{h}$ is classified via Linear$(64\!\to\!1)$. In all fusion models, the rPPG waveform is encoded by Linear$(160\!\to\!64)$, batch normalization, ReLU, and dropout (10K parameters), and the DCT vector by the DCT-only encoder; the 1D ResNet is used only in the rPPG-only baseline. We also report rPPG-only with this linear encoder.

\textbf{Concat fusion.} Both 64-dim encodings are concatenated to 128-dim, classified via Linear$(128\!\to\!64)\!\to\!$ReLU$\!\to\!$Dropout$\!\to\!$Linear$(64\!\to\!1)$.

\textbf{Cross-attention fusion.} rPPG is patched into 20 query tokens of dim 64 via Conv1d$(1,64,k=8,s=8)$, DCT into 8 key-value tokens. Multi-head attention \cite{vaswani2017attention} (4 heads) attends rPPG to DCT; outputs are mean-pooled and classified by a $64\!\to\!32\!\to\!1$ MLP.

\begin{figure}[!t]
\centering
\includegraphics[width=\columnwidth, trim=0 15 10 10,clip]{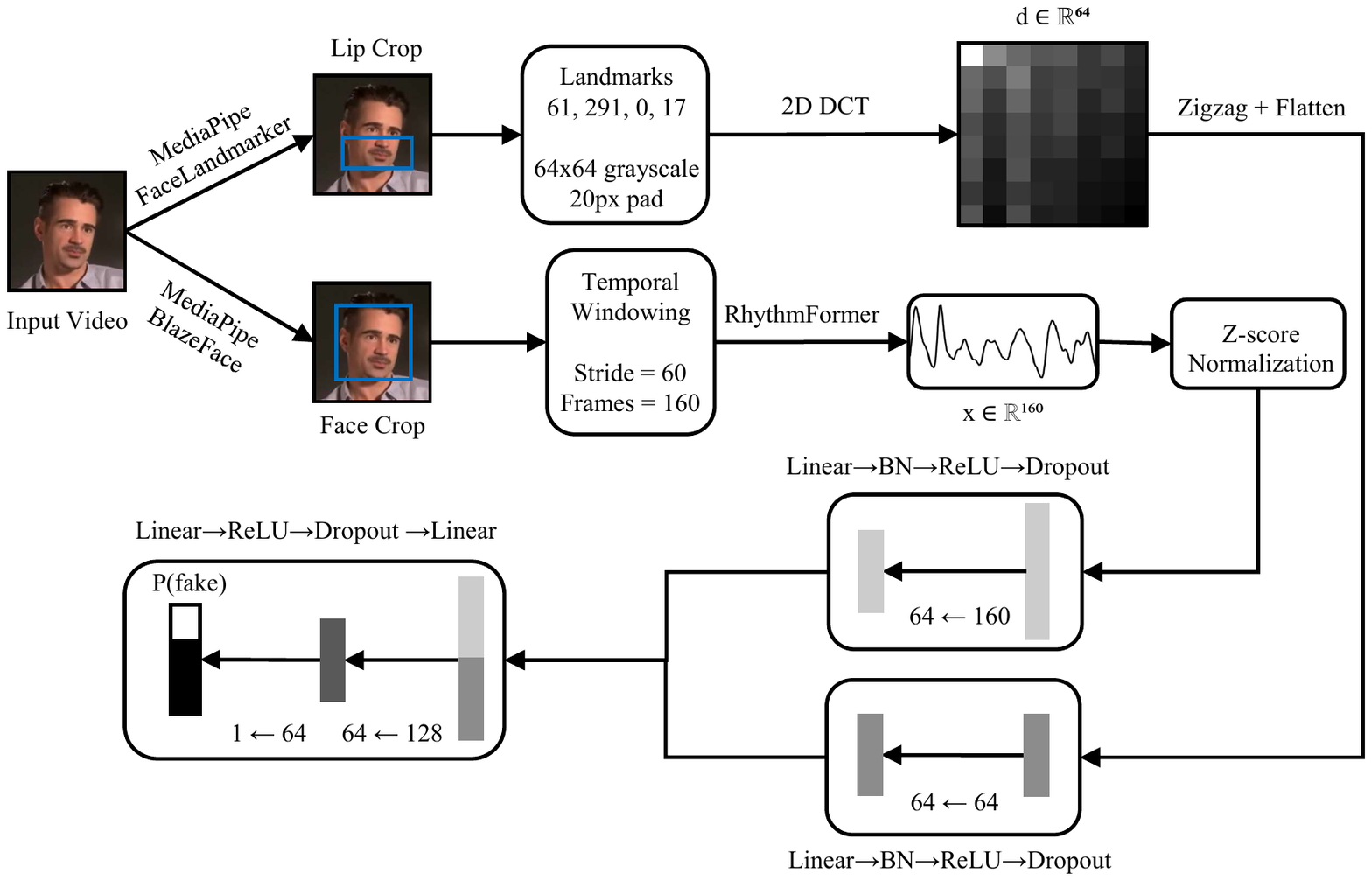}
\caption{Proposed rPPG$+$DCT concat fusion pipeline. The rPPG branch (bottom) yields a z-scored waveform $\mathbf{x}\in\mathbb{R}^{160}$; the frequency branch (top) yields lip-region DCT coefficients $\mathbf{d}\in\mathbb{R}^{64}$. Both are projected to 64-dim, concatenated, and classified. Bar heights indicate dimensionality.}
\label{fig:pipeline}
\end{figure}

All models use AdamW (lr$=10^{-3}$, wd$=5\times10^{-4}$), dropout 0.3 (0.5 for the ResNet), batch size 64, 30 epochs, weighted binary cross-entropy with $w_+ = n_-/n_+$. The full pipeline is illustrated in Fig.~\ref{fig:pipeline}.

\subsection{Evaluation Protocol}
\label{sec:protocol}
We evaluate on the TF subset of Celeb-DF++ \cite{li2025celebdf}: 590 real videos from 59 celebrity identities, of which 579 yield a usable waveform (2,371 real windows), and 17,497 TF-forged videos (2,500 randomly sampled per method; three EchoMimic videos lack features). This fake set is smaller than the 20,279 used in \cite{harraq2027paper1}, so our rPPG-only results differ slightly from that paper (Combined AUC 0.824 vs 0.826). The 59 identities are split into train (41), validation (9), and test (9) with zero overlap. Models train for a fixed 30 epochs without early stopping or checkpoint selection, and we report window-level AUC on the 18 validation and test identities combined (696 real windows, 5,553 fake videos), following \cite{harraq2027paper1}. All results are mean $\pm$ std over 5 seeds.

\section{Results}
\label{sec:results}

\subsection{Fusion Architecture Comparison}

Table~\ref{tab:fusion} reports per-method AUC. Concat is the strongest fusion model (Combined 0.891), with Gated close behind (0.880). SadTalker is the only method where Concat trails rPPG-only (0.913 vs 0.946). Part of this gap reflects the linear rPPG encoder used inside fusion models, which alone reaches 0.927 on SadTalker, but Concat also trails that baseline; using the ResNet as Concat's rPPG encoder raises SadTalker to 0.932 while lowering IP-LAP to 0.828. The mean per-sample gate weight $\alpha$ exceeds 0.5 only on SadTalker (0.611) and falls to 0.10--0.14 on FLOAT and EchoMimic; it weights the two encodings but does not directly measure each branch's contribution to accuracy. Cross-attention (0.810) trails DCT-only overall and matches it on SadTalker (0.706 vs 0.712): its output is formed only from attended DCT values, with rPPG shaping the attention weights.

\begin{table}[t]
\caption{Per-method AUC. Rows: one model per method. Combined: one model trained on all seven methods, pooled AUC. rPPG: 1D ResNet; rPPG$_\text{lin}$: the linear encoder used inside fusion models. 5 seeds, std $\le$ 0.010.}
\label{tab:fusion}
\centering
\footnotesize
\renewcommand{\arraystretch}{1.1}
\begin{tabular}{lcccccc}
\toprule
\textbf{Method} & \textbf{rPPG} & \textbf{rPPG$_\text{lin}$} & \textbf{DCT} & \textbf{Gated} & \textbf{Con.} & \textbf{CA} \\
\midrule
AniTalker      & 0.923 & 0.903 & 0.953 & 0.971 & \textbf{0.978} & 0.945 \\
EchoMimic      & 0.878 & 0.824 & 0.990 & 0.992 & \textbf{0.993} & 0.984 \\
EDTalk         & 0.956 & 0.946 & 0.965 & 0.985 & \textbf{0.990} & 0.956 \\
FLOAT          & 0.824 & 0.769 & 0.984 & 0.984 & \textbf{0.987} & 0.979 \\
IP-LAP         & 0.698 & 0.728 & 0.792 & 0.837 & \textbf{0.854} & 0.767 \\
Real3DPortrait & 0.988 & 0.990 & 0.989 & \textbf{0.998} & \textbf{0.998} & 0.987 \\
SadTalker      & \textbf{0.946} & 0.927 & 0.712 & 0.913 & 0.913 & 0.706 \\
\midrule
Combined       & 0.824 & 0.804 & 0.827 & 0.880 & \textbf{0.891} & 0.810 \\
\bottomrule
\end{tabular}
\end{table}

\subsection{Leave-One-Generator-Out Generalization}
\label{sec:logo}

Table~\ref{tab:lomo} reports leave-one-generator-out (LOGO) results: each method is held out entirely while models train on the other six, with reals and fakes split by identity as in Sec.~\ref{sec:protocol}. The cues split by held-out method: the rPPG-derived branch transfers better to AniTalker, EDTalk, and SadTalker, where DCT falls to 0.622--0.671, and DCT transfers better to EchoMimic, FLOAT, and Real3DPortrait; held-out IP-LAP is near chance for both. Concat has the highest mean (0.798, vs 0.752 and 0.729), but falls below rPPG-only on the three methods where DCT transfers poorly, by up to 0.138 on SadTalker. An oracle choosing the better single cue per held-out method would average 0.818, so static fusion only partly exploits the complementarity. The rPPG-only LOGO mean is close to that of \cite{harraq2027paper1} (0.754).

\begin{table}[t]
\caption{Leave-one-generator-out AUC. Train on six methods, test on the held-out one; identity-disjoint, 5 seeds. std $\le$ 0.017, except rPPG on Real3DPortrait (0.035). Mean: unweighted over held-out methods.}
\label{tab:lomo}
\centering
\footnotesize
\renewcommand{\arraystretch}{1.1}
\begin{tabular}{lccc}
\toprule
\textbf{Held-out} & \textbf{rPPG} & \textbf{DCT} & \textbf{Concat} \\
\midrule
AniTalker      & \textbf{0.839} & 0.657 & 0.816 \\
EchoMimic      & 0.650 & 0.906 & \textbf{0.908} \\
EDTalk         & \textbf{0.848} & 0.671 & 0.768 \\
FLOAT          & 0.771 & 0.896 & \textbf{0.910} \\
IP-LAP         & 0.554 & 0.534 & \textbf{0.563} \\
Real3DPortrait & 0.736 & 0.815 & \textbf{0.893} \\
SadTalker      & \textbf{0.865} & 0.622 & 0.727 \\
\midrule
Mean           & 0.752 & 0.729 & \textbf{0.798} \\
\bottomrule
\end{tabular}
\end{table}

\subsection{Lip-Region vs Full-Face DCT}

Table~\ref{tab:dct} compares lip-region and full-face DCT. Lip-region DCT is better on six of seven methods; on AniTalker the two differ by 0.001. The largest drop from lip to full face is on IP-LAP (DCT-only: 0.792 to 0.641; Concat: 0.854 to 0.755). Since the full-face variant summarizes a larger region with the same 64 coefficients, this comparison is consistent with, but does not isolate, a mouth-localized cue.

\begin{table}[t]
\caption{Lip-Region vs Full-Face DCT. 5 seeds, std $\le$ 0.010. Lip-region outperforms full-face on six of seven methods for DCT-only.}
\label{tab:dct}
\centering
\footnotesize
\renewcommand{\arraystretch}{1.1}
\begin{tabular}{lcccc}
\toprule
 & \multicolumn{2}{c}{\textbf{DCT only}} &
   \multicolumn{2}{c}{\textbf{Concat}} \\
\cmidrule(lr){2-3}\cmidrule(lr){4-5}
\textbf{Method} & \textbf{Lip} & \textbf{Face} &
                  \textbf{Lip} & \textbf{Face} \\
\midrule
AniTalker      & 0.953 & \textbf{0.954} & \textbf{0.978} & 0.975 \\
EchoMimic      & \textbf{0.990} & 0.979 & \textbf{0.993} & 0.983 \\
EDTalk         & \textbf{0.965} & 0.945 & \textbf{0.990} & 0.985 \\
FLOAT          & \textbf{0.984} & 0.969 & \textbf{0.987} & 0.974 \\
IP-LAP         & \textbf{0.792} & 0.641 & \textbf{0.854} & 0.755 \\
Real3DPortrait & \textbf{0.989} & 0.966 & \textbf{0.998} & 0.997 \\
SadTalker      & \textbf{0.712} & 0.623 & \textbf{0.913} & 0.904 \\
\midrule
Combined       & \textbf{0.827} & 0.771 & \textbf{0.891} & 0.870 \\
\bottomrule
\end{tabular}
\end{table}

\subsection{Per-Method Complementarity}

\begin{figure}[t]
\centering
\includegraphics[width=\columnwidth]{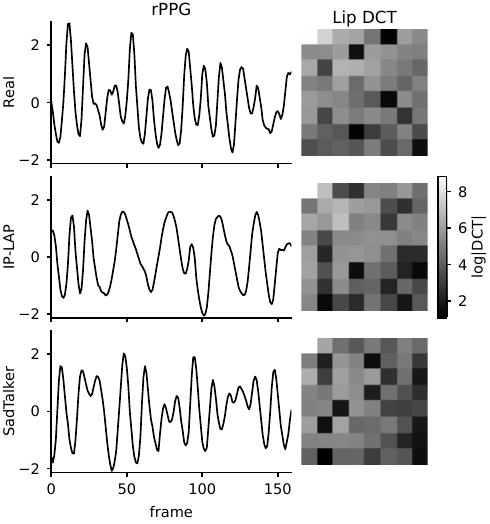}
\caption{Representative rPPG-derived waveforms (left) and lip-region DCT log-magnitude maps (right, shared color scale) for a real video, an IP-LAP fake, and a SadTalker fake of one identity. Per-method AUC: rPPG 0.698 (IP-LAP) vs 0.946 (SadTalker); DCT 0.792 vs 0.712. Illustrative, not quantitative evidence.}
\label{fig:waveforms}
\end{figure}

Fig.~\ref{fig:waveforms} shows representative examples for IP-LAP and SadTalker, where the two cues diverge most.

\begin{figure}[t]
\centering
\includegraphics[width=\columnwidth]{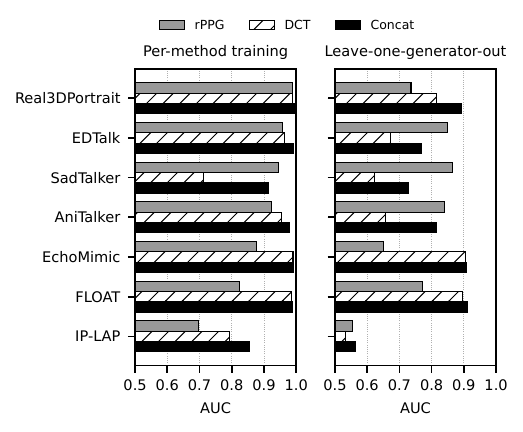}
\caption{Per-method AUC for the rPPG-derived branch (1D ResNet), lip-region DCT, and Concat. Left: per-method training. Right: leave-one-generator-out. Axis starts at chance.}
\label{fig:inverted}
\end{figure}

Fig.~\ref{fig:inverted} summarizes both settings. In-domain, the rPPG-derived branch substantially outperforms DCT only on SadTalker (0.946 vs 0.712), while DCT leads on FLOAT and EchoMimic by more than 0.1; IP-LAP is hardest for both cues, and there fusion yields its largest gain over the stronger unimodal baseline ($+$0.062). The right panel shows the LOGO split of Sec.~\ref{sec:logo}.

\section{Discussion}
\label{sec:discussion}

\subsection{Method-Dependent Complementarity}

Neither cue is strongest across methods. In-domain, DCT matches or exceeds the rPPG-derived 1D ResNet on six of seven methods but falls to 0.712 on SadTalker, where the rPPG-derived branch reaches 0.946, and the mean gate weight favors rPPG only there. Under LOGO, the complementarity is sharper: each cue transfers clearly better to a different three held-out methods, and Concat trails the better single cue on three of them. Fusion that selects a cue per input or per inferred generator may recover more of this gap; we evaluate only Concat under LOGO. Synthesis strategy is a tempting explanation, but seven methods cannot separate strategy from other generator properties, and \cite{harraq2027paper1} found that detectability and transferability do not follow a single strategy variable. IP-LAP by design re-renders the lower face of a template video, and its Celeb-DF++ fakes are named by a real template clip; if upper-face footage is retained, this would explain why rPPG-derived detection is weakest there. The rPPG-derived branch may also reflect temporal appearance statistics rather than cardiac content \cite{harraq2027paper1}.

\subsection{Limitations}
Evaluation covers only the TF subset of Celeb-DF++, with real video from 59 identities. As in \cite{harraq2027paper1}, real videos are windowed across their full length while each fake contributes one window, and real and generated videos differ in frame rate (30 vs 24--25 fps; extraction uses native frame rates, while the RhythmFormer checkpoint was trained on 30 fps video); both asymmetries may contribute to the rPPG-derived signal. DCT features may likewise reflect resolution or encoding differences between real and generated video. Method labels follow the release's directory structure. rPPG extraction is sensitive to skin tone and lighting \cite{nowara2020meta}, and we do not stratify results by demographic group.

\section{Conclusion}
\label{sec:conclusion}

In-domain, lip-region DCT matches or exceeds the rPPG-derived 1D ResNet on every method except SadTalker, and Concat fusion reaches AUC 0.891. Under identity-disjoint leave-one-generator-out evaluation, each cue transfers clearly better to three held-out methods while IP-LAP remains near chance for both; Concat averages 0.798 but trails the better single cue on three, motivating fusion that selects a cue per input. Future work should test these patterns on face-swap and face-reenactment forgeries.

\section{Acknowledgments}
AI tools assisted with portions of the code, manuscript editing, and an initial draft of the references; the authors verified all content.

\bibliographystyle{IEEEbib}
\bibliography{main}

\end{document}